%% file: pcgm_arxiv.tex
\documentclass[runningheads]{llncs}
\usepackage[a4paper, margin=1in]{geometry}
\usepackage{cite}
\usepackage{float}
\usepackage{amsmath}
\usepackage{amssymb}
\usepackage{amsfonts}
\usepackage{mathtools}
\usepackage{algorithmic}
\usepackage{graphicx}
\usepackage{textcomp}
\usepackage{booktabs}
\usepackage{multirow}
\usepackage{color}
\usepackage{rotating}
\usepackage{tabularray}
\usepackage{array}
\usepackage{url}
\usepackage{hyperref}

\begin{document}
\title{Integrating Anatomical Priors into a Causal Diffusion Model}
\author{
Binxu Li\inst{1}$^{*}$ \and
Wei Peng\inst{2}$^{*}$ \and
Mingjie Li\inst{2} \and
Ehsan Adeli\inst{2} \and
Kilian M. Pohl\inst{2}\,$^{\dagger}$
}

\authorrunning{B. Li et al.}

\institute{
Department of Electrical Engineering, Stanford University, Stanford, CA, USA \\
\email{andy0207@stanford.edu}
\and
Department of Psychiatry and Behavioral Sciences, Stanford University, Stanford, CA, USA \\
\email{\{wepeng, lmj695, eadeli, kpohl\}@stanford.edu} \\
$^*$Equal contribution. \quad $^\dagger$Corresponding author.
}

% \thanks{This work was partly supported by the National Institute of Health (AA021697, DA057567, AA010723, AA05965, and AA017347), and by the Stanford University Human-Centered Artificial Intelligence.}
% \thanks{NCANDA data collection and distribution were supported by NIH funding AA021681, AA021690, AA021691, AA021692, AA021695, AA021696, AA021697, AG089169. They are made publicly accessible via \url{https://nda.nih.gov/edit_collection.html?id=4513}}
% \thanks{Binxu Li is with the Department of Electrical Engineering, Stanford University (e-mail: andy0207@stanford.edu). Drs. Peng, Li, Adeli, and Pohl are with the Department of Psychiatry and Behavioral Sciences,
% Stanford University (e-mail: \{wepeng, lmj695, eadeli, kpohl\}@stanford.edu). By courtesy, Dr. Adeli is with the Department of Computer Science, Stanford University and Dr. Pohl with the Department of Electrical Engineering, Stanford University.}}

\newcommand{\upcausalencoder}{Causal Encoder}
\newcommand{\causalencoder}{causal encoder}
\newcommand{\diffusiondecoder}{3D diffusion decoder}
\newcommand{\updiffusiondecoder}{3D Diffusion Decoder}
\newcommand{\modulesecond}{Counterfactual Mask Generator}
\newcommand{\abbrmodulesecond}{CMG}
\newcommand{\modulethird}{Mask Guided Diffusion}
\newcommand{\abbrmodulethird}{MGD}
\newcommand{\diff}{Counterfactual Diffusion Module}
\newcommand{\abbrdiff}{CDM}

\maketitle

\begin{abstract}

3D brain MRI studies often examine subtle morphometric differences between cohorts that are hard to detect visually. Given the high cost of MRI acquisition, these studies could greatly benefit from image syntheses, particularly counterfactual image generation, as seen in other domains, such as computer vision. However, counterfactual models struggle to produce anatomically plausible MRIs due to the lack of explicit inductive biases to preserve fine-grained anatomical details. This shortcoming arises from the training of the models aiming to optimize for the overall appearance of the images (e.g., via cross-entropy) rather than preserving subtle, yet medically relevant, local variations across subjects. To preserve subtle variations, we propose to explicitly integrate anatomical constraints on a voxel-level as prior into a generative diffusion framework. Called Probabilistic Causal Graph Model (PCGM), the approach captures anatomical constraints via a probabilistic graph module and translates those constraints into 
spatial binary masks of regions where subtle variations occur. The masks (encoded by a 3D extension of ControlNet) constrain a novel counterfactual denoising UNet, whose encodings are then transferred into high-quality brain MRIs via our \diffusiondecoder. Extensive experiments on multiple datasets demonstrate that PCGM generates structural brain MRIs of higher quality than several baseline approaches. Furthermore, we show for the first time that brain measurements extracted from counterfactuals (generated by PCGM) replicate the subtle effects of a disease on cortical brain regions previously reported in the neuroscience literature. This achievement is an important milestone in the use of synthetic MRIs in studies investigating subtle morphological differences.


\end{abstract}

\section{Introduction}
\label{sec:introduction}

The generation of high-fidelity structural brain MRIs is increasingly important in medical imaging research and clinical practice, as structural brain MRIs are indispensable for investigating neurodevelopment\cite{mri_s1}, monitoring disease progression\cite{mri_s2}, and developing AI-assisted diagnostic tools\cite{mri_s3}. However, the acquisition of 3D MRI scans is limited by factors such as scanner availability, lengthy scan times, and high costs, resulting in relatively small and fragmented datasets~\cite{GAN_gen, mrigen}. These brain MRI studies thus could greatly benefit from synthetically generated MRIs as high-quality synthetic data can augment limited datasets and support AI-driven diagnosis and research~\cite{jindal2024genAI,wang2025self,wang2025toward,bluethgen2024vision,tudosiu2024realistic}.
 
While progress in 2D image generation has been substantial \cite{rombach2022ldm, sd3, dit}, extending these models to generate anatomically plausible 3D MRIs remains challenging \cite{wu2024eval, singh2021medical}. In addition to having to account for the high voxel dimensionality of MRIs ($>$5M)  coupled with confining training to relatively small datasets ($<$ 100K), existing models prioritize reconstruction of global appearances (by optimizing cross-entropy) rather than focusing on preserving fine-grained morphology on a local level that is important for studying many neuropsychiatric conditions~\cite{wu2024eval}. For instance, mild cognitive impairment \cite{MCI}, HIV \cite{hivehsan}, and alcohol use disorder (AUD)  \cite{sullivan2018role} are associated with subtle morphological changes in several cortical regions that are not visible to the naked eye. These subtle changes are not captured by state-of-the-art (SOTA) models \cite{sun2025eval}, which highlights the need for more models that are tailored towards the specific needs of brain MRI studies. 

\begin{figure*}[!t]
    \centering
    \includegraphics[width=1\textwidth]{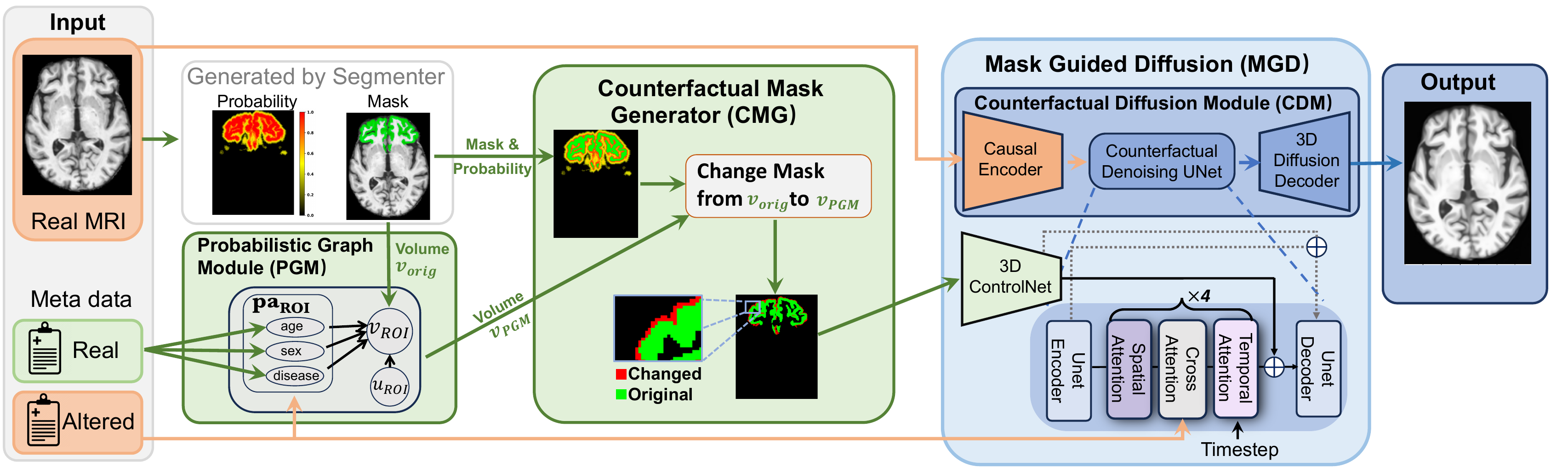}
    \caption{\textbf{The framework of our Probabilistic Causal Graph Model (PCGM).} Known causal relationships between metadata (e.g., age, sex, disease label) and the volumes of brain regions of interest (ROIs) are encoded via the \textbf{Probabilistic Graph Module (PGM)}. Given the altered volume of an ROI based on modified metadata (intervention),  the \textbf{\modulesecond\ (\abbrmodulesecond)}  modifies the mask of the ROI extracted from the original MRI to match the new volume score. Based on the original MRI, the metadata, and the counterfactual mask encoded by \textbf{3D ControlNet}, the \textbf{\modulethird\ (\abbrmodulethird)} module generates the corresponding MRI counterfactuals using our proposed \textbf{\diff\ (\abbrdiff)}.}
    \label{fig:method}
    % \vspace{-0.6cm}
\end{figure*}

This need becomes even more critical in the context of counterfactual generation, which aims to provide a response to the fundamental question of how a brain changes if a condition of a subject is altered (such as transforming the MRI of a healthy control into one if the person were diagnosed with AUD) \cite{cf_mri}. Learning to capture these intra-subject changes frequently requires training on longitudinal data, which are generally of an even smaller sample size than cross-sectional data used for training unconditional generative modes. One possible solution is to train models on MRIs from multiple datasets~\cite{wu2024eval,puglisi2025brain}. However, the resulting sample size would still be less than 1K for most diseases, i.e., too small to robustly train entirely data-driven methods on. An alternative to data pooling could be causal generative models~\cite{fabio2023cfmodel,yeganeh2025latent, peng2024latent}, which first extract key brain measurements from the MRIs and then model their dependencies with respect to metadata using a causal graph. However, these methods so far have only been designed to capture coarse, visible effects (e.g., ventricular enlargement in Alzheimer’s disease)\cite{nestor2008ventricular}. To generate counterfactual structural MRIs that account for subtle changes in the brain, we propose a diffusion-based counterfactual model that explicitly integrates anatomical knowledge on a voxel-level into the generation process.

Called the Probabilistic Causal Graph Model (PCGM; see Fig.~\ref{fig:method}), the approach first learns to encode the interactions among subject-level factors (e.g., sex, age) and brain regional measurements using a probabilistic graph model (PGM). The ROI measurements (e.g., volumes of cortical regions) generated by a PGM are then turned into binary masks. 
% The resulting ROI masks are encoded by 3D ControlNet (our 3D extension of \cite{controlnet}) whose output together with altered metadata confines our counterfactual denoising UNet. 
The resulting ROI masks are encoded by a 3D ControlNet (our 3D extension of~\cite{controlnet}), whose outputs, together with the altered metadata, provide conditioning for our Counterfactual Denoising UNet.
% that is coupled with a causal VAE–latent diffusion architecture \cite{causalvae}. 
Instead of directly decoding the latent features produced by that UNet with a standard VAE decoder (as done by most LDM architectures \cite{rombach2022ldm, monaildm}), we pretrain a dedicated \diffusiondecoder to reconstruct the final 3D brain MRI, which ensures that the output is of high image quality and anatomical plausibility.

We first train our approach on 3954 t1w MRIs of controls from ADNI~\cite{ADNI} and NCANDA~\cite{NCANDA}. Evaluation is based on a matched, hold-out set of 400 subjects from both data sets and an in-house dataset consisting of 199 controls, 222 participants diagnosed with AUD, and 41 individuals with HIV \cite{hivdata}. Experimental results demonstrate that our model consistently produces MRIs with higher image quality and stronger anatomical plausibility than all baseline models. Furthermore, only the counterfactual MRIs generated by PCGM on the in-house dataset yield cortical brain measurements that replicate the subtle effects of AUD on the brain published on the same data set by~\cite{sullivan2018role}. To our knowledge, PCGM is the first approach to achieve this significant milestone in generating MRIs useful for neuroscience studies. 

\section{Related Works}

\subsection{MRI Synthesis}
Traditional methods for MRI synthesis have relied on deforming real MRIs rather than directly generating MRI intensities~\cite {zitova2003image}. These methods map real MRIs to a template, modify the deformation field or the template itself, and then map the results back to create a synthesized MRI~\cite{zitova2003image,freeborough1996accurate,maintz1998survey,hill2001medical}. More recently, deep learning has enabled direct intensity-based synthesis by encoding the intensity distributions of MRIs~\cite{chung2022score}. Approaches within this framework include variational autoencoders (VAEs) ~\cite{vae_mri} and generative adversarial networks (GANs)~\cite{GAN_gen}, with GANs achieving notable success. 

The two types of GAN-based approaches are image-to-image transformations and unconditional generation. Image-to-image transformations create new MRIs from existing ones and have been applied to cross-modality synthesis~\cite{MultiContrastGAN2019}, counterfactual generation~\cite{shin2018medical}, and tumor simulation~\cite{yu20183d}. However, these models often require large, well-curated datasets, and their ability to increase data diversity is limited~\cite{karras2019style,xing2021cycle}. Unconditional generation, on the other hand, encodes the underlying distribution of the dataset to produce entirely new samples starting from random noise~\cite{bermudez2018learning,han2018gan}. While recent advances by methods such as $\alpha$-WGAN~\cite{kwon2019generation} and 4D-DANI-Net~\cite{ravi2022degenerative} have improved the quality of the generated MRIs, GANs still face challenges with model collapse, high memory requirements, and unstable training, limiting their ability to generate realistic, high-quality 3D MRIs~\cite{peng2024brainsyn}.

An alternative approach is the denoising diffusion probabilistic model (DDPM)~\cite{ho2020ddpm,dhariwal2021diffusion}, which synthesizes images by gradually transforming a Gaussian distribution into a target distribution through a Markov chain process. While computationally intensive, recent improvements adopting non-Markovian processes have made DDPMs more efficient, enabling applications in medical image analysis, including anomaly detection~\cite{la2022anatomically}, segmentation~\cite{wolleb2022diffusion}, and MRI acceleration~\cite{chung2022score}. Extending DDPMs to 3D MRI synthesis typically involves adapting 2D operations to 3D~\cite{dorjsembe2022three}, though this can still be computationally prohibitive and may not reach the same quality as GAN-based MRI synthesis~\cite{peng2023cDPM}. Additionally, extending diffusion models to longitudinal MRI synthesis remains challenging due to high computational demands~\cite{yoon2023sadm}.

MedGen3D~\cite{han2023medgen3d} addresses these limitations by synthesizing MRIs slice-by-slice, conditioning each slice on prior ones to reduce resource requirements. However, this approach can result in artifacts, such as inconsistent intensities between slices~\cite {peng2023cDPM}. Recently, BrainSyn~\cite{peng2024brainsyn} was proposed to generate high-resolution, subject-agnostic 3D brain volumes, achieving anatomically accurate structures compared to previous methods. Nevertheless, a human expert was reliably distinguishing synthetic from real MRIs as the synthetic MRIs contained subtle artifacts (such as always showing a vessel at exactly the same location in the brain).

\subsection{Counterfactual MRI Generation}

Counterfactual MRI synthesis aims to introduce anatomically plausible variations in an MRI~\cite{fabio2023cfmodel}.  For example, models like CounterSynth~\cite{pombo2023equitable} employ conditional generative frameworks to produce realistic, diffeomorphic deformations based on counterfactual labels, thereby generating anatomically accurate variations that reflect specified conditions. Furthermore, approaches~\cite{yeganeh2025latent,wang2025toward,puglisi2025brain} have further advanced counterfactual MRI synthesis by introducing metadata into unified representations and training text-guided generative models to synthesize 2D and 3D brain images conditioned on descriptive prompts. However, these methods are limited to diffeomorphic transformations, restricting their capacity to enhance data diversity. Some GAN-based approaches attempt counterfactual synthesis via image-to-image transformations~\cite{thiagarajan2022training}, yet they similarly rely on available training pairs, leading to constrained diversity in generated samples. Recently, \cite{fabio2023cfmodel} proposed a high-fidelity counterfactual model for lung CTs and brain MRIs. Due to the complexity of 3D counterfactuals, their work focused on 2D slices and on clearly visible changes related to aging, such as ventricular enlargement with age. In contrast, our 3D counterfactual model targets nuanced structural changes that are often observed in psychiatric studies, thereby advancing the capacity for subtle, anatomically plausible modifications of MRIs.

\section{Methodology}

We now describe in further detail PCGM (Fig.~\ref{fig:method}), our approach for generating accurate counterfactuals of MRIs. Given a t1w brain MRI, the approach first extracts a volume score, mask, and probability map of each region of interest (ROI) via $\text{SynthSeg}^+$~\cite{billot_robust_2023}. Next, the volume scores together with the original metadata (such as age and diagnoses) are fed into a probabilistic graph module (PGM) in order to update the scores according to the modified metadata. Given the `intervened' scores and the original masks, the \modulesecond\ (\abbrmodulesecond) produces new binary masks for each ROI. Finally, the modified masks guide the generation of the counterfactual MRI by the \modulethird\ (\abbrmodulethird) module. We now describe these three modules in further detail and end with how to use PCGM to generate counterfactual MRIs. 

\subsection{Probabilistic Graph Module (PGM)}
PGM models known causal relationships between metadata (e.g., age, sex, and diagnosis) and ROI volume scores using a {deep structural causal model (SCM)}~\cite{pawlowski2020deep}. In SCM, the causal relationships are encoded in a directed graph consisting of observed (endogenous) variables $V := \{v_1, \dots, v_{N}\}$ (i.e., meta data and volumes of ROIs), which are generated by causal mechanisms $F := \{f_1, \dots, f_{N}\}$ applied to independent (exogenous) noise variables $U := \{u_1, \dots, u_N\}$ and the parent variables $\mathbf{pa}_k$ (i.e., none for metadata and metadata for the volume scores) so that  
\begin{equation}
v_k := f_k(\mathbf{pa}_k, u_k).
\label{eqn:v}
\end{equation}
$f_k$ is parameterized via a normalizing flow~\cite{rezende2015variational} so that $f_k$ is invertible, which allows us to estimate the exogenous noise as
\begin{equation}
u_k = f_k^{-1}(v_k; \mathbf{pa}_k).
\label{eqn:u}
\end{equation}
Now, let $\mathbf{\widehat{pa}}_k$ be the counterfactual values of the parents of $v_k$ after the intervention, then (following \cite{de2023high}) one can compute its counterfactual value $\widehat{v}_{k}$ by replacing $u_k$ in Eq \ref{eqn:v} with its definition (Eq \ref{eqn:u}):
\[
\widehat{v}_{k} \coloneqq f_{k}(f_{k}^{-1}(v_{k}; \mathbf{pa}_{k}); \mathbf{\widehat{pa}}_k).
\]

The estimation of counterfactuals follows the standard three-step SCM process of \textbf{abduction, action,} and \textbf{prediction}. \textbf{Abduction} infers the posterior distribution $P(U \mid V)$ from the observations, representing the latent noise consistent with the current metadata. Next, \textbf{action} applies an intervention using the do-operator, e.g., $do(v_k = c)$, which sets an endogenous (parent) variable to a fixed value (such as setting a specific diagnosis). Finally, \textbf{prediction} uses the modified model together with the abducted noise to propagate changes and compute the new values of all downstream variables. In this way, the PGM can simulate how hypothetical changes (e.g., altering a diagnosis) would influence ROI volumes in a principled, probabilistic manner.

\subsection{\modulesecond~(\abbrmodulesecond)}
Based on the counterfactual ROI volume scores generated by PGM, our approach now modifies the corresponding mask of each region (Fig.~\ref{fig:method}). Here, we describe modifying the masks of cortical regions, as subtle differences in those regions are often reported in psychiatric studies, such as for AUD in \cite{sullivan2018role}, whose findings we aim to replicate later. However, the approach described below can be easily extended to other brain regions impacted by other diseases. 

To model the subtle impact of psychiatric diseases on $M$ cortical ROIs, one needs to know that small changes to an ROI also change the neighboring CSF but not the white matter~\cite{jernigan1991reduced}. In other words, changing the volume of an ROI needs to result in a change of the boundary between the ROI and CSF, but not the white matter. Thus, one cannot simply erode or dilate masks in accordance with the countfactual volume as this would imply changing both CSF and white matter, i.e., be unrealistic. 
For an arbitrary cortical ROI $k\in \left\{ 1, \ldots M\right\}$, we instead first identify the boundary between CSF and the ROI based on the segmentation provided by SynthSeg$^+$. Around this boundary, we define a voxel-level probability map \(P_{k}\) from the probability maps generated by SynthSeg$^+$. Thus \(P_{k}(l=i|x)\) is the probability that label $i \in \left\{ \mbox{white matter}, \mbox{CSF}, \mbox{ROI}_K\right\}$ is assigned to voxel $x$. 

Now lets assume that the mask of the ROI $k$ needs to be \textbf{increased}, i.e., its original volume \(v_{\text{orig}}\) (in voxels) is smaller than the counterfactual volume \(v_{\text{PGM}}\) determined by the PGM.  
To increase the mask by  \(d:=v_{\text{PGM}} - v_{\text{orig}}\) voxels, we simulate 'increasing' the voxel-level probabilities \(P_{k}(l=k)\) by a scalar $\alpha > 1$ for that ROI until $d$ voxels are added. Doing so is equivalent to ranking the boundary voxels $x$ labeled as CSF according to the difference in probabilities \mbox{\(P_{k}(l=\mbox{ROI}_K|x) - P_{k}(l=\mbox{CSF}|x)\)} and then increasing the mask by the top $d$ voxels. 

In case the mask of the ROI needs to be \textbf{decreased}, we cannot simply decrease the probabilities of the ROI as voxels around the boundary could then be assigned to white matter (or another ROI) instead of CSF. Instead, we repeat the same procedure as above, but now increase the probability of CSF.

\subsection{\modulethird~(\abbrmodulethird)}
% Given the mask for each cortex, we now generate the counterfactual based on  3D causal VAE-based diffusion module and a 3D ControlNet~\cite{zhang2023addingconditionalcontroltexttoimage}.
Guided by the counterfactual masks produced by the \abbrmodulesecond, our \modulethird\ (\abbrmodulethird) module produces counterfactual MRI based on three key components: (1) a \causalencoder ~that embeds MRIs into a latent space; (2) a novel counterfactual denoising UNet that conditions the latent encoding of MRIs with respect to counterfactual metadata and mask; and (3) a \diffusiondecoder ~that transforms the MRI encodings into high-resolution MRIs. Each of these components is now described in further detail:

\subsubsection{\upcausalencoder} Recognizing the importance of capturing subtle details in generating MRIs, we rely on a continuous VAE by adapting the causal architecture~\cite{causalvae} (often used on videos\cite{bain2022frozentimejointvideo}) to 3D MRIs. Specifically, in each batch, our model views each slice of an MRI as a static frame and learns to encode those 2D frames while simultaneously learning cross-slice (or frame) consistency to properly account for constraints within the entire 3D volume. To minimize biases in the encoding towards the acquisition plane of the slices (i.e., axial, coronal, or sagittal), we randomly permute the order of planes for each MRI before we start training the encoder. Once training is completed, each MRI is encoded in a latent space. 

\subsubsection{Counterfactual Denoising UNet} As in MedSyn~\cite{xu2024medsyn}, we rely on a fully 3D CNN encoder-decoder UNet backbone augmented with self-attention layers. The UNet encoder is followed by four attention blocks consisting of three attention layers: (1) a 3D volumetric attention layer, which extends MedSyn’s 2D spatial attention to operate directly on volumetric patches so that it can better capture anatomical context in 3D; (2) a cross-attention mechanism that integrates counterfactual metadata with MRI features (replacing the text-based conditioning in MedSyn); and (3) a temporal attention aligned with the diffusion timestep to model the dynamics of the denoising process.

In addition to metadata-based conditioning, we further incorporate the counterfactual mask to directly guide the diffusion process on a voxel-level. Specifically, we first encode the 3D mask via 3D ControlNet, which consists of replacing conv (designed for 2D images) with conv3D in ControlNet~\cite{zhang2023adding}. Using the same layer structure as the UNet encoder, we combine the features extracted by an encoder layer of 3D ControlNet with those extracted by the corresponding UNet layer through the skip connections and with the output of the attention block for the final layer. Guided by the diffusion noise schedule of the diffusion module, this design progressively incorporates the mask as the image denoises, ensuring the output adheres to the voxel-level condition defined by the mask at each denoising step.
.

\begin{figure*}[!t]
    \centering
    \includegraphics[width=1\linewidth]{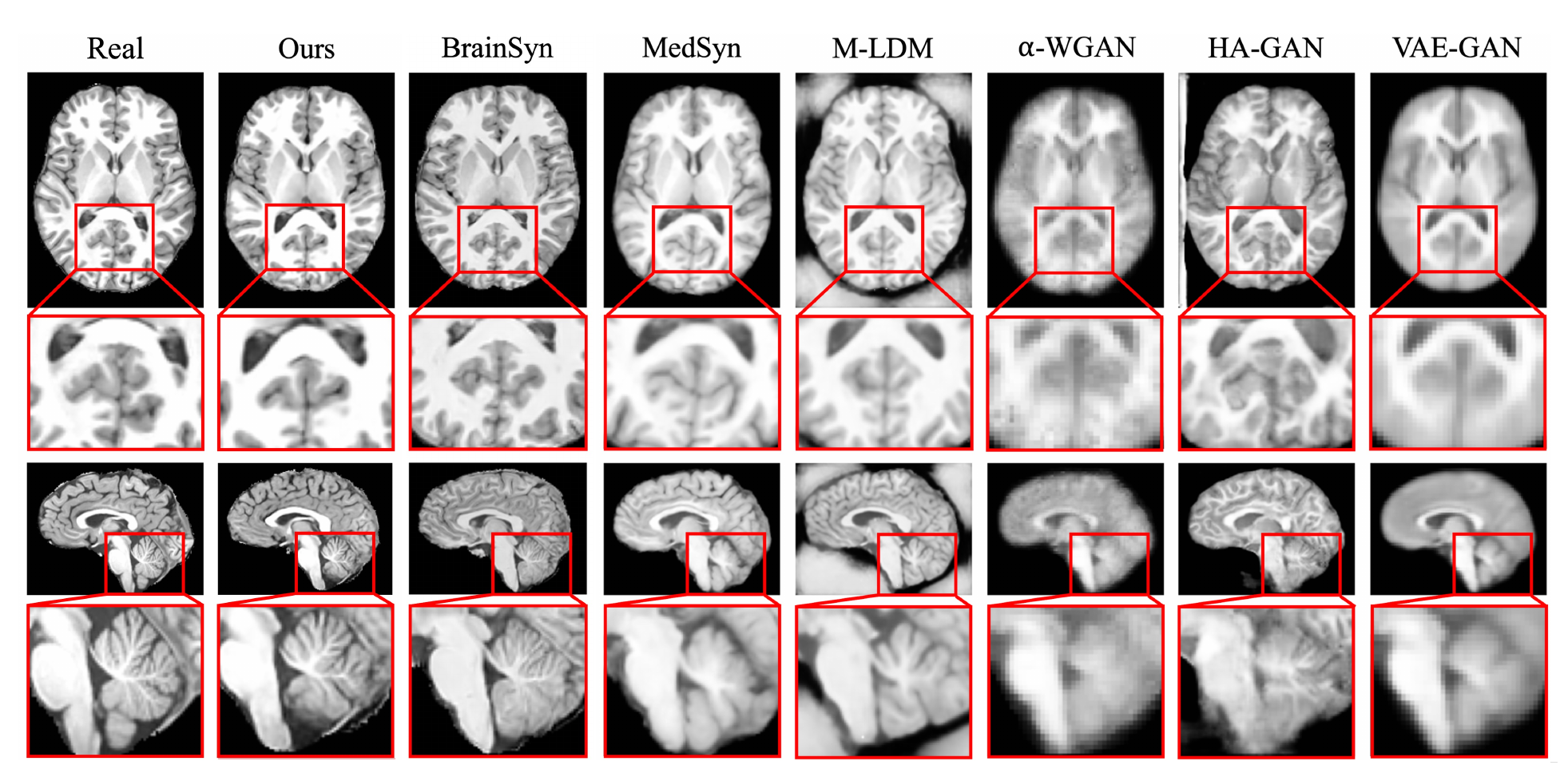}
    \caption{\textbf{Real and synthetic MRIs created by generative approaches.} Given the real MRI, the figure displays the most similar synthetic MRI generated by each method. Each MRI is displayed from the axial (first row) and sagittal view (third row). Compared to baseline approaches, the image contrast and visibility of anatomical structures in the MRI generated by our method most closely match the real MRI, especially for the cerebellum (fourth row). }
    \label{sotaComp}
\end{figure*}

\subsubsection{\updiffusiondecoder} 

The diffusion decoder proposed in ~\cite{diffvae} reconstructed high-fidelity 2D images by leveraging a diffusion model. Unlike the feedforward upsampling decoders used in VAEs~\cite{Kingma2013VAE}, this diffusion decoder employs an iterative denoising process that progressively refines noisy samples into realistic images conditioned on latent representations. Inspired by their success, we design the first 3D CNN-based diffusion decoder that transforms the counterfactual latent representation $z_c$ (generated by the Counterfactual Denoising UNet) into high-fidelity counterfactual MRIs. Unlike in \cite{diffvae}, we first upsample $z_c$ to $z_c’$ so that it matches the spatial resolution of the target MRI $x_0$. Conditioned on $z_c’$, the diffusion decoder then denoises a sequence of noisy samples in order to approximate the data distribution $p(x_0|z_c’)$. Through this iterative refinement, the decoder not only preserves subtle anatomical structures but also enhances sample fidelity and diversity, which are critical in medical imaging where small variations may carry important clinical meaning.

\subsection{Counterfactual Generation}
To generate a counterfactual, we first perform the intervention on the metadata and apply the PGM to obtain the corresponding values of the leaf nodes (i.e., volumes of the ROIs). For each ROI, the \abbrmodulesecond\ modifies its mask according to the volume scores. Next, DDIM inversion~\cite{mokady2022inv} is applied to the 3D causal encoding of the original MRI, which produces the reverse-generation sequence that aligns with the original metadata. By selecting an intermediate latent state from this sequence as the initialization point for denoising, the model enables counterfactual generation of the MRI under given altered metadata conditions. Finally, conditioned on both the counterfactual mask and the modified metadata, we apply the DDIM inversion and regeneration process to synthesize disease-specific counterfactual MRIs that reflect the intervention while preserving overall anatomical consistency.

\input{exp_new_arxiv}

\section{Conclusion}
%% limitaion
This work introduces Probabilistic Causal Graph Model (PGCM), a novel causal diffusion model for generating counterfactual MRI of high anatomical plausibility. PGCM consists of the Probabilistic Graph Module (PGM) for capturing known causal relationships between metadata and ROIs, the \modulesecond\ (\abbrmodulesecond) for modifying the mask of ROI to match the volume given by PGM, and the \modulethird\ (\abbrmodulethird) for high-fidelity 3D MRI synthesis based on the counterfactual denoising UNet. Even with limited training data, this approach accurately captures both broad and subtle changes in MRIs linked to aging and alcohol use disorder (AUD) as revealed by our experiments. With respect to AUD, the counterfactual MRIs generated by our approach were able to replicate published findings on subtle cortical changes, which is an important milestone for advancing disease modeling and synthetic data generation for neuroscience research.

\section{Acknowledgements}
This work was partly supported by the National Institute of Health (AA021697, DA057567, AA010723, AA05965, and AA017347), and by the Stanford University Human-Centered Artificial Intelligence. NCANDA data collection and distribution were supported by NIH funding AA021681, AA021690, AA021691, AA021692, AA021695, AA021696, AA021697, AG089169. They are made publicly accessible via \url{https://nda.nih.gov/edit_collection.html?id=4513}.

{
    \bibliographystyle{ieee}

    \bibliography{main}
}

\end{document}

%% file: exp_new_arxiv.tex
\section{Experiments}
\label{sec:exp}

\begin{table}[t]
\caption{\textbf{Comparison between generative models.} The absolute relative difference in MS-SSIM between synthetic and real MRIs (MS-SSIM: 0.767), and the FID and the MMD scores in both image and feature spaces (using 3D ResNet 101(R101) and ResNet 50(R50)) produced by 7 generative approaches. The top scores are in bold and the second-best scores are underlined. Our method produces the best scores in 4 categories and the second-best score in the remaining two, underlying its overall superiority over the other baselines. \label{table:SOTA}}
\resizebox{0.97\columnwidth}{!}{
% \small
    \begin{tabular}{l|cc|cc|c|c}
    % \toprule
    \textbf{\begin{tabular}[c|]{@{}c@{}}Model\end{tabular}} & \multicolumn{2}{c|}{\textbf{\begin{tabular}[c|]{@{}c@{}}ResNet-R101 \\ FID($\downarrow$) \quad MMD($\downarrow$) \end{tabular}}}  & \multicolumn{2}{c|}{\textbf{\begin{tabular}[c]{@{}c@{}} ResNet-R50 \\ FID($\downarrow$)\quad MMD($\downarrow$) \end{tabular}}}& \textbf{\begin{tabular}[c]{@{}c@{}}Image \\ MMD($\downarrow$)\end{tabular}} & \textbf{\begin{tabular}[c]{@{}c@{}}MS-\\ SSIM($\downarrow$)\end{tabular}} \\ \hline 
    \textbf{VAE-GAN~\cite{vqvae2017}} & 0.032   & 0.020     & 0.400 & 0.210  &  \textbf{142069} & 5.96\% \\ \hline
    \textbf{$\alpha$-WGAN~\cite{ferreira2022alphaGAN}}  & 0.032  & 0.020  & 0.496   &  0.258 &214578 & 3.28\% \\ \hline
    \textbf{HA-GAN~\cite{sun2022hagan}}  & 0.036 & 0.020      & 0.088 & 0.056 & 767583 & 13.25\% \\ \hline 
        \textbf{M-LDM~\cite{pinaya2022brain}}  & 0.320  & 0.160     & 1.910 & 0.960 &  3432589 & 35.21\%  \\ \hline
    \textbf{MedSyn~\cite{xu2024medsyn}} &0.014  & 0.012    & 0.048  & 0.036  &  245896 & \underline{1.86\%}  \\ \hline
     \textbf{BrainSyn~\cite{peng2024brainsyn}} & \underline{0.005} & \underline{0.007}   & \underline{0.022}  & \textbf{0.022} &  300841  & 2.99\%  \\ \hline
     \textbf{Ours} & \textbf{0.001} & \textbf{0.006}    & \textbf{0.011} & \underline{0.030}  &  \underline{208150}  & \textbf{1.06}\% 
    % \bottomrule
    \end{tabular}}
\end{table}

\subsection{Experimental Setup}

To document the strengths and weaknesses of our approach, we systematically evaluate our method in three stages. First, we focus on assessing the quality of MRIs generated just by the unconditional diffusion model of our method (i.e., by omitting metadata and the mask; the blue components in Fig. \ref{fig:method}). Next, we add meta-data (but still omit the mask generated by the \abbrmodulesecond) to the model to create longitudinal counterfactuals (i.e., counterfactuals with respect age; blue and orange components in Fig \ref{fig:method}) and assess the accuracy of the generated counterfactuals with respect to the ventricles, i.e., a structure that visually increases in size with age. Finally, we test the complete approach PCGM (green, orange and blue components in Fig \ref{fig:method}) in replicating the subtle effects of alcohol use disorder (AUD) on cortical structures as reported in \cite{sullivan2018role}. To ease readability, we now first describe the experimental setup and findings of the first two tasks. The resulting model will then be the base of PCGM, which will be tested in the third experiment.

\subsubsection{Datasets}  
The first two tasks are based on 1273 t1w MRIs from all 380 controls of the Alzheimer's Disease Neuroimaging Initiative (ADNI, baseline age: 75.5 $\pm$ 6.1, female ratio: 50.72\%, Data Releases: ADNI 1, 2, 3 and GO)~\cite{petersen2010alzheimer} and 3081 t1w MRI from 767 healthy participants of the National Consortium on Alcohol and Neurodevelopment in Adolescence (NCANDA, baseline age: 16.1 $\pm$ 2.5, female ratio: 51.53\%, Data Releases: NCANDA\_PUBLIC\_6Y\_STRUCTURAL\_V01)~\cite{NCANDA} that passed through our processing pipeline~\cite{peng2023cDPM}. We split the joint data set into 667 subjects (consisting of 3954 MRIs) for training and the remaining 400 subjects (132 from ADNI, 268 from NCANDA) for testing so that the two subsets were matched with respect to sex and age.  

While the first task tests the generative approach on the baseline MRIs of all 400 subjects, the test set for the second task is further reduced to the 7 longitudinal MRIs from the NCANDA data set with at least 5 visits and the 23 longitudinal MRIs from ADNI. The test set of the second task is complemented with an out-of-sample data set of longitudinal MRIs (consisting of two visits) of 41 participants with HIV (age: 53.3 $\pm$ 7.8, female ratio: 36.5\%) of the SRI-Stanford study (PI: Sullivan, Pfefferaum)~\cite{hivdata}. 
% Each subject had two t1w MRIs acquired an average of 71 months apart from each other. 

All T1-weighted MRIs were preprocessed following~\cite{peng2023cDPM}, including denoising, skull stripping, registration, and intensity normalization. In addition, each MRI was segmented using SynthSeg$^+$~\cite{billot_robust_2023}. 

\subsubsection{Implementation Details}
All experiments were run on a single NVIDIA A100 GPU (80GB). For the first two tasks, we trained the foundational components of our model, i.e., the \upcausalencoder, counterfactual denoising UNet, and the \updiffusiondecoder ~(blue and orange components in Fig. \ref{fig:method}). The \upcausalencoder ~of the \abbrmodulethird ~was initialized from a pre-trained OpenSora CausalVAE~\cite{causalvae} and fine-tuned for 40,000 iterations using a batch size of 1, a learning rate of 1e$^{-5}$, and cropped MRIs to dimension 101x104x104. Next, we trained the counterfactual denoising UNet conditioned on z-scored metadata (i.e., age and sex) and the \updiffusiondecoder ~separately for 40K iterations with a learning rate of 8e$^{-6}$. To further improve the 'alignment'  between (altered) metadata and generated MRIs, we relied on classifier-free guidance \cite{cfg}.  Finally, we generated 400 MRIs for Task 1 from random noise without any metadata conditioning. For Task 2, we generated counterfactuals for each test subject by obtaining an intermediate latent sequence through DDIM inversion of the baseline MRI. We then adjusted the age of the model to the age of the subject at subsequent visits and performed DDIM generation to produce counterfactual MRIs. For each MRI, we measured the ventricular volume using SynthSeg$^+$. We focused on the ventricles as these are visibly increasing in size as individuals get older (i.e., we do not require the mask encoding brain regions).

\subsection{Task 1: General Brain Synthesis}
The 400 MRIs produced by our generative approach were compared against those generated by three baseline GAN approaches (i.e,  VAE-GAN~\cite{vqvae2017}, $\alpha$-WGAN~\cite{ferreira2022alphaGAN}, and HA-GAN~\cite{sun2022hagan}) and three diffusion models (i.e.,  M-LDM~\cite{pinaya2022brain}, MedSyn~\cite{xu2024medsyn}, and BrainSyn~\cite{peng2024brainsyn}). Each method was trained and evaluated using the same experimental setup as our method.

\subsubsection{Qualitative Results}
As shown in Fig.~\ref{sotaComp}, the brain MRIs generated by the four diffusion models are of superior visual quality compared to GAN-based approaches. Of the diffusion models, the MRIs of M-LDM and MedSyn are still quite fuzzy, i.e., gray matter boundaries are not clearly defined. Brainsyn and our approach produce visually very nice MRIs. However, the MRI generated by our approach is the only one showing clear gray matter boundaries within the cerebellum (last row in Fig.~\ref{sotaComp}).

\begin{figure}[!t]
    \centering
    \includegraphics[width=0.9\linewidth]{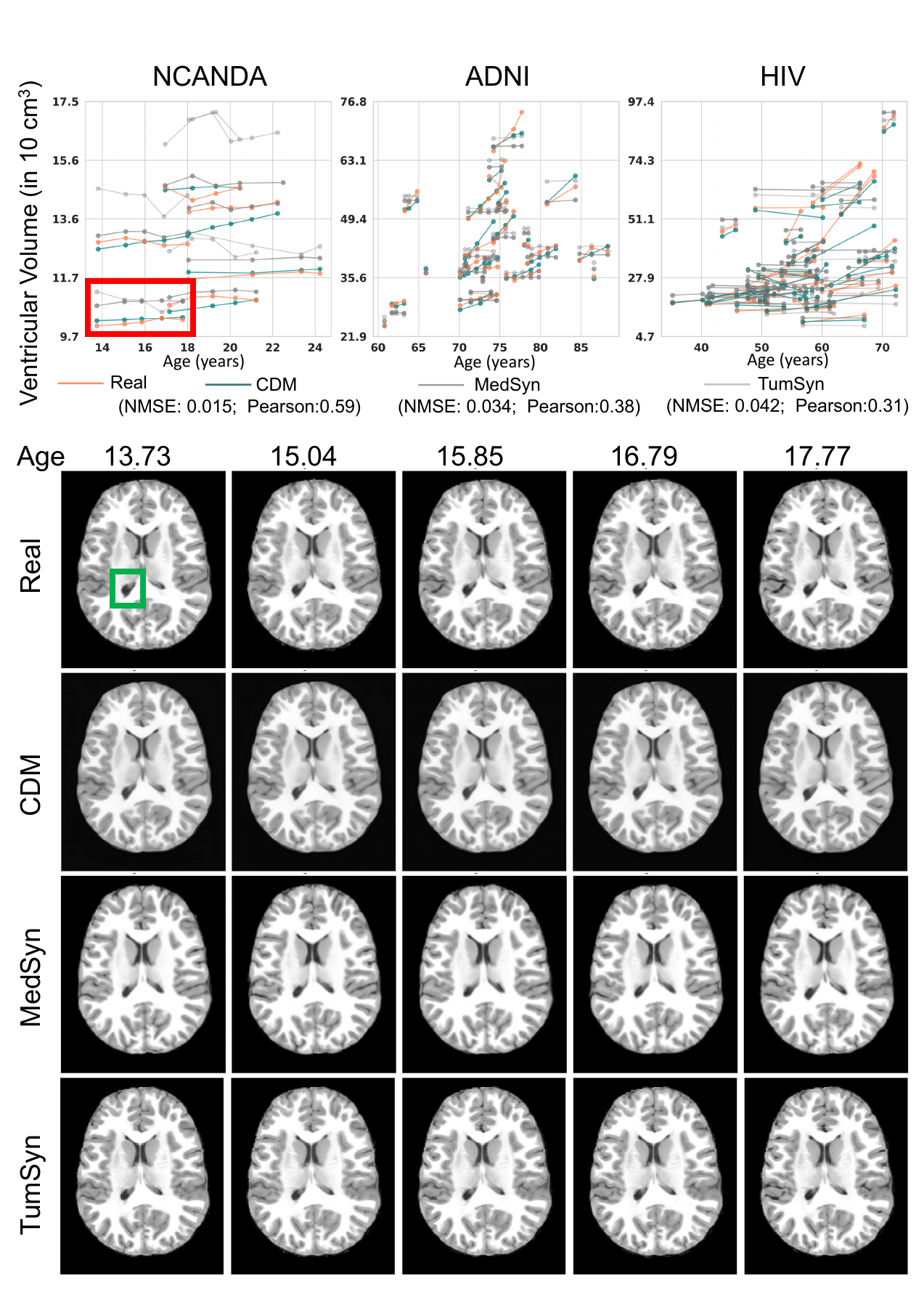}
    \caption{\textbf{Age trajectory modeling.} Each trajectory in the plot represents the volume of the ventricle of one subject measured at different ages. The subjects were participants of NCANDA (ages 14 - 26 years), ADNI (ages 60 - 90 years), or diagnosed with HIV (ages 33 - 74 years). Among the three methods, the trajectories extracted from the counterfactual MRIs produced by our proposed \abbrdiff ~align the best (i.e., lowest NMSE and highest Peatson score) with those of the real MRIs. This is also confirmed when visually comparing the longitudinal MRIs of the trajectories outlined in red. Visually the largest differences are shown in the ventricle region outlined in green.
    \label{fig:agetrajectory}}
\end{figure}

\subsubsection{Quantitative Results}
This observation is also supported by the quantitative evaluation summarized in Table \ref{table:SOTA}. Specifically, the 400 MRIs produced by each approach were assessed using the traditional imaging metrics Fréchet Inception Distance (FID) \cite{fid} and Maximum Mean Discrepancy (MMD), which were calculated in both feature and image spaces. To assess the MRIs in feature space, we employed the pre-trained 3D medical networks 3D ResNet-101 and ResNet-50 as in~\cite{sun2022hagan}. In addition, we recorded their MS-SSIM \cite{msssim} score and computed the $\%$ difference with respect to the MS-SSIM recorded on the 400 real MRIs that were omitted from the training set. Table \ref{table:SOTA} reveals that our approach achieves the top or second-best score in all 6 categories. VAE-GAN achieves the top score for the MMD score computed in the image space as its MRIs are overly smooth and have a fuzzy appearance, which reduces pixel-level discrepancies. BrainSyn and our method achieve the best MMD scores based on the ResNet-50 and ResNet-101 encodings. However, with respect to the two metrics on the imaging space (i.e., voxel space), our method clearly outperforms BrainSyn indicating that subtleties of MRIs are probably not well captured by the metrics based on the Res-Net encodings, which were originally derived for natural images. Overall, this comparison highlights that the generative model of our counterfactual approach synthesizes MRIs of superior image quality than all basline methods. 

Since traditional image-level metrics may not accurately capture anatomical plausibility, we followed the evaluation pipeline proposed in~\cite{wu2024eval}. Specifically, we applied FreeSurfer~\cite{freesurfer} to each real and synthetic MRI to record the volume of 34 cortical brain regions as defined by the Desikan-Killiany atlas (Aparc) and Freesurfer's image quality control (QC) score. Only models whose QC scores were at least as high as those recorded on the real MRIs were then included in the comparison, which were MedSyn, BrainSynth, and our method. For each cortical region and method, we then computed the absolute Cohen’s d score ($|d|$) between the volumetric distributions of the synthetic vs. real MRIs. According to~\cite{peng2024brainsyn}, $|d| < 0.2$ is seen as a small effect (i.e., good alignment with the real data), while $0.2 < |d| < 0.5$ indicates medium effects, and $|d| > 0.5$ suggests a large effect implying greater deviation from the real data. 

\begin{figure}[t]
    \centering
    \includegraphics[width=0.9\linewidth]{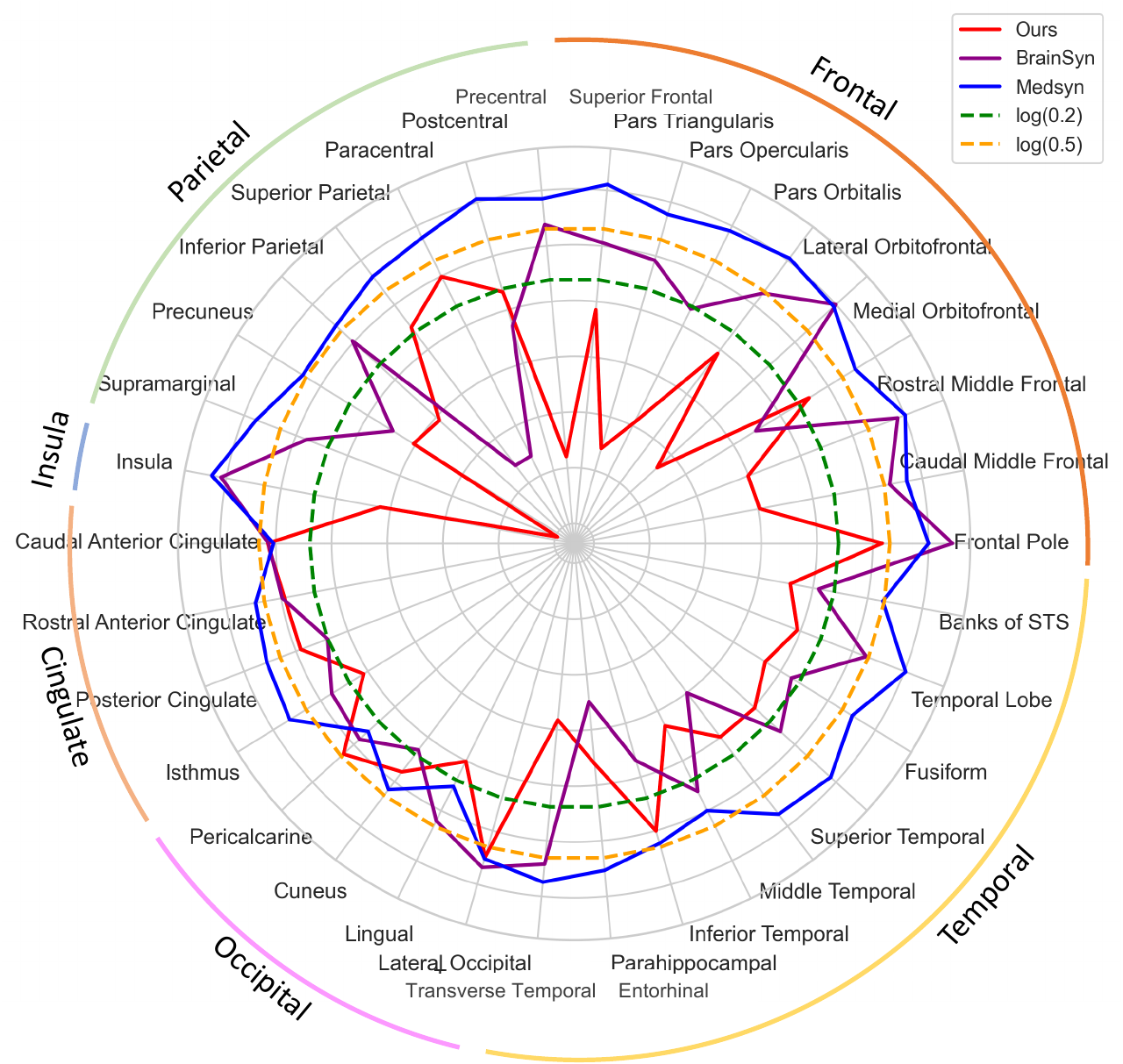}
    \caption{\textbf{Anatomical Plausibility}. Displayed are the natural logarithm of the absolute Cohen's d ($|d|$) for 32 regions and three methods. $|d| \textless$ 0.2 is viewed as a small effect, 0.2 $\textless |d| \textless$ 0.5 indicates a medium effect, and $|d|>$ 0.5 corresponds to a large effect. Our method records the smallest d-score for the majority of regions, indicating the highest overall anatomical plausibility among the 3 methods. Note, the plot connects the scores between ROIs simply to ease comparison.}
    \label{fig:anatomy}
\end{figure}

According to Fig.~\ref{fig:anatomy}, the worst performing approach among the three is Medsyn, whose $|d|$ is always higher than that of our method with the exception of parahippocampus (MedSyn: 0.38, Our: 0.43) and caudal anterior cingulate (MedSyn: 0.26, Our: 0.46). While BrainSyn is generally better than Medsyn, it is somewhat unstable as it produces the worst overall $|d|$ with 1.54 in the frontal pole and exceeds the critical 0.5 threshold for 24\% of the regions. In comparison, our approach exceeds that threshold only for one region (i.e., lateral occipital lobe: $|d|$=0.53), is well aligned with volume scores from the real MRI for 62.8 $\%$ of the regions, and produces the lowest $|d|$ across all three methods in 67.6 $\%$ of the regions. In summary, the anatomical plausibility of MRIs produced by our approach is generally high and superior to these baseline methods. 

\subsubsection{User Study}

We further assessed the ability of our model to generate 3D MRIs of high realism by repeating the user study of the second-best approach BrainSyn \cite{peng2024brainsyn}. Specifically, we randomly selected 50 real MRIs from the test set. For each MRI we then identified the most similar synthetic MRI generated by our method. These 100 MRIs (50 real and 50 synthetic) were then randomly mixed and independently evaluated by the same three experts as in prior work \cite{peng2024brainsyn}. Each expert has over 20 years of experience in reading MRIs. They were asked to label each MRI as either real or synthetic. The overall average classification accuracy across the three experts was 46.0\% (see Table \ref{tab:user_study}). This indicates that the experts were not able to distinguish real from synthetic MRIs, which was not the case for BrainSyn \cite{peng2024brainsyn} (overall accuracy was 70.7\%). 

\begin{table}[t]
\caption{Accuracy of identifying real and synthetic MRIs by experts.}
\centering
\resizebox{0.6\linewidth}{!}{
\begin{tabular}{lccc}
\toprule
        & Real & Synthetic & Overall accuracy \\
\midrule
Expert 1 & 46.0\% & 54.0\% & 50.0\% \\
Expert 2 & 50.0\% & 44.0\% & 47.0\% \\
Expert 3 & 42.0\% & 40.0\% & 41.0\% \\
\midrule
Average  & 46.0\% & 46.0\% & 46.0\% \\
\bottomrule
\end{tabular}
}
\label{tab:user_study}
\end{table}

%
% -----------------------------------------
%

\subsection{Task 2: Aging Counterfactual Generation}
 
We now review the findings of our simple counterfactual approach (blue and orange components in Fig. \ref{fig:method}), which we refer to as \diff ~(\abbrdiff). For comparison, we repeat the analysis for MedSyn (using the same experimental setup as for \abbrdiff) and TUMSyn~\cite{wang2025toward}. For TUMSyn, we used its officially released, pretrained weights. Those weights were generated from 31,407 3D images across 7 modalities collected from 13 studies, which included 500 T1w MRIS from ADNI (a subset of our training set), 2864 from ABCD (age range: 9--19)~\cite{abcd}, 1000 from UKBioBank (age range: 44--82)~\cite{sudlow2015ukbiobank}, 1100 from HCPY (22--35)~\cite{van2012HCP}. Counterfactual brains were generated by modifying the age specification in the text prompt of TUMSyn. Like the other two approaches, we record the ventricular volume of the longitudinal MRIs generated by TUMSyn.

\begin{table*}[!t]
    \centering
    \caption{\textbf{Replicating AUD findings} Inline with \cite{sullivan2018role}, cortical measurements extraced from the counterfactual MRIs of our method (PCGM) successfully reproduces the significant group differences (p-value < 0.00833, i.e., p-value of 0.05 after Bonferroni correction) between control and subjects diagnosed with alcohol use disorder (AUD) for five of the six brain regions (whether real or synthetic). In contrast, no significant differences are correctly reported within the same group (i.e., between real and synthetic controls or between real and synthetic AUDs), indicating that our counterfactuals preserve group identity while capturing meaningful disease-related differences~\cite{sullivan2018role}. This is not the case for \abbrdiff~(i.e., PCGM without the mask) and TumSyn, which fail to reproduce the findings of \cite{sullivan2018role}. This indicates that simply conditioning on label signals is insufficient for the model to capture the complex structural changes associated with disease.}
    \resizebox{1\columnwidth}{!}{%
    {%
    \begin{tabular}{l|c|cc|cc|cc}
     & \multirow{2}{*}{\textbf{Region}} & \multicolumn{2}{c|}{\textbf{Counterfactual Only}} & \multicolumn{4}{c}{\textbf{Original MRI vs. Counterfactual}} \\
     & & \multicolumn{2}{c|}{Control vs. AUD} & Control vs. AUD & AUD vs. Control & Control vs. Control & AUD vs. AUD \\
    \hline
    \multirow{7}{*}{\rotatebox{90}{PCGM}} & Frontal   & \multicolumn{2}{c|}{\textcolor{green}{\textbf{\textless 0.0001}}}  & \textbf{\textcolor{green}{\textless 0.0001}} & \textcolor{green}{\textbf{\textless 0.0001}}  & \textcolor{green}{0.4242} & \textcolor{green}{0.1928} \\
    & Insula               & \multicolumn{2}{c|}{\textcolor{green}{\textbf{0.0013}}}           & \textcolor{green}{\textbf{\textless 0.0001}}          & \textcolor{green}{\textbf{0.0078}}            & \textcolor{green}{0.3312}   & \textcolor{green}{0.4976} \\
    & Parietal   & \multicolumn{2}{c|}{\textcolor{green}{\textbf{0.0015}}}   & \textcolor{green}{\textbf{\textless 0.0001}} & \textcolor{green}{\textbf{\textless 0.0001}}           & \textcolor{green}{0.7331}  & \textcolor{green}{0.9256} \\
    & Cingulate   & \multicolumn{2}{c|}{\textcolor{green}{\textbf{0.0022}}}  & \textcolor{green}{\textbf{0.0023}}     &   \textcolor{green}{\textbf{0.0064}}     & \textcolor{green}{0.3018}  & \textcolor{green}{0.3377} \\
    & Temporal    & \multicolumn{2}{c|}{\textcolor{green}{\textbf{\textless 0.0001}}}           & \textcolor{green}{\textbf{\textless 0.0001}} & \textcolor{green}{\textbf{\textless 0.0001}}           & \textcolor{green}{0.1038}  & \textcolor{green}{0.5042} \\
    & Occipital &  \multicolumn{2}{c|}{\textcolor{green}{0.8242}}         & \textcolor{green}{0.5428}         & \textcolor{green}{0.9891}          & \textcolor{green}{0.9778}   & \textcolor{green}{0.5341} \\
    \cline{2-8}
    & \# Correct      &  \multicolumn{2}{c|}{6}         & 6       & 6           & 6  & 6\\
    \hline \hline
    \multirow{7}{*}{\rotatebox{90}{\abbrdiff}} & Frontal  &  \multicolumn{2}{c|}{\textcolor{green}{\textbf{0.0022}}}  & \textcolor{red}{0.1231}  & \textcolor{red}{0.3315} & \textcolor{green}{0.4566} & \textcolor{green}{0.1979}  \\
    & Insula   &  \multicolumn{2}{c|}{\textcolor{red}{0.0122}}  & \textcolor{red}{0.3813}  & \textcolor{red}{0.1291} & \textcolor{green}{0.1083} & \textcolor{green}{0.0568} \\
    & Parietal   &  \multicolumn{2}{c|}{\textcolor{green}{\textbf{0.0012}}}  & \textcolor{red}{0.1757} & \textcolor{red}{0.2977} & \textcolor{green}{0.1557}  & \textcolor{green}{0.3338} \\
    & Cingulate  &  \multicolumn{2}{c|}{\textcolor{green}{\textbf{0.0059}}}  & \textcolor{red}{0.4667} & \textcolor{red}{0.3648}  & \textcolor{green}{0.1077}  & \textcolor{green}{0.2828} \\
    & Temporal    &  \multicolumn{2}{c|}{\textcolor{red}{0.0093}}         & \textcolor{red}{0.2118} & \textcolor{red}{0.8977}           & \textcolor{green}{0.0243}  & \textcolor{green}{0.0338} \\
    & Occipital     &  \multicolumn{2}{c|}{\textcolor{green}{0.5241}}         & \textcolor{green}{0.4318}        & \textcolor{green}{0.3277}           & \textcolor{green}{0.0891}  & \textcolor{green}{0.1233} \\
    \cline{2-8}
    & \# Correct         &  \multicolumn{2}{c|}{4}         & 1       & 1           & 6 & 6 \\
    \hline \hline
    \multirow{7}{*}{\rotatebox{90}{TumSyn}} & Frontal  &  \multicolumn{2}{c|}{\textcolor{green}{\textbf{0.0033}}}  & \textcolor{red}{0.9837}  & \textcolor{red}{0.6342} & \textcolor{red}{\textbf{0.0007}} & \textcolor{red}{\textbf{0.0009}}  \\
    & Insula   &  \multicolumn{2}{c|}{\textcolor{green}{\textbf{0.0028}}}  & \textcolor{red}{0.9321}  & \textcolor{red}{0.8632} & \textcolor{red}{\textbf{0.0017}} & \textcolor{red}{\textbf{0.0019}} \\
    & Parietal  &  \multicolumn{2}{c|}{\textcolor{red}{0.0542}}  & \textcolor{red}{0.9318} & \textcolor{red}{0.7302} & \textcolor{red}{\textbf{0.0005}}  & \textcolor{red}{\textbf{0.0024}} \\
    & Cingulate  &  \multicolumn{2}{c|}{\textcolor{red}{0.0132}}  & \textcolor{red}{0.8136} & \textcolor{red}{0.9218}  & \textcolor{red}{\textbf{0.0008}}  & \textcolor{red}{\textbf{\textless 0.0001}} \\
    & Temporal         &  \multicolumn{2}{c|}{\textcolor{red}{0.0242}}         & \textcolor{red}{0.4367} & \textcolor{red}{0.3917}           & \textcolor{red}{\textbf{\textless 0.0001}}  & \textcolor{red}{\textbf{0.0009}} \\
    & Occipital        &  \multicolumn{2}{c|}{\textcolor{green}{0.5518}}         & \textcolor{green}{0.8128}        & \textcolor{green}{0.7477}           & \textcolor{green}{0.7439}  & \textcolor{green}{0.8255} \\
    \cline{2-8} 
    & \# Correct         &  \multicolumn{2}{c|}{3}         & 1       & 1           & 1 & 1 \\
    \end{tabular}
    }% end local arraystretch group
    }% end resizebox
    \vspace{-0.6cm}
    \label{tab:ad_pvalues_combined}
\end{table*}

% \subsubsection{Results} 
As shown in Fig.~\ref{fig:agetrajectory}, the slope of the trajectories based on our synthetically generated MRIs align well with the real data, confirming our model's ability to learn and replicate aging effects in the brain. The same conclusion cannot be drawn for the other two counterfactual methods, whose ventricular volumes sometimes largely deviate from the real cases. This difference is also visible in the example longitudinal MRI shown in that figure. 

To confirm this qualitative assessment, we compare each synthetic longitudinal MRI to the corresponding real sequence by computing the Pearson correlation coefficient and average normalized mean square error (NMSE) of the volume scores at the subject level and perform paired t-tests on NMSE scores between our method and other baseline methods. As shown Fig.~\ref{fig:agetrajectory}, our method achieves the best average Pearson coefficient of 0.59, compared to MedSyn (0.38) and TumSyn (0.31). It also achieves a significantly smaller NMSE value (p-values $\textless$ 0.001 compared to scores of the two methods) demonstrating that our method more accurately captures the aging process of the ventricles.

% ------------------------------
% 
\subsection{Task 3: Disease Modeling}

The third task tests the abilities of our proposed counterfactual approach PCGM to replicate the findings with respect to AUD on the brain originally reported in \cite{sullivan2018role}. The dataset in \cite{sullivan2018role} consists of 826 t1W MRIs acquired of 222 participants diagnosed with AUD (baseline age: 48.05 $\pm$ 10.33; female ratio:  31.13\%) and 199 age-matched healthy controls (age: 47.21 $\pm$ 12.64; female ratio: 42.89\%).  Based on the encoder, diffusion model, and decoder trained for Tasks 1\&2, we used the pretrained U-Net of the \abbrdiff ~to initialize the weights of the 3D ControlNet, which is then trained on the training data of Task 1 $\&$ 2. We then used  5-fold cross-validation to train and test the remaining components of our approach. Specifically, for each test fold of the out-of-sample AUD data set, we then used the remaining data to train the PGM module, which creates a causal graph relating scalar variables (i.e.,  age, sex, diagnosis) to the volume of cortical regions (Frontal, Parietal, Insula, Cingulate and Temporal). Training of the PGM model ran for 50,000 iterations with a batch size of 64 and a learning rate of 1e$^{-5}$. For each test subject, we then computed its counterfactual by computing the cortical volume under diagnostic interventions. Together with the mask and probability map, these changes were then the input to the \abbrmodulesecond, which produced a modified mask for each cortical region. The modified mask and the real MRI were then the input to the \abbrmodulethird, which produced the counterfactual. 
% Training .... 

To replicate the findings in \cite{sullivan2012using}, we first regressed out the supratentorial volume from each of the cortical volume measurements from each of the cortical volume scores (Frontal, Parietal, Insula, Cingulate, Occipital and Temporal). For each ROI, we computed its average value across all visits for each subject. Across subjects, we then identified significant differences (p-value $\textless$ 0.00833, i.e., p-value of 0.05 after Bonferroni correction for six regions) between control vs. AUD on the real data. 

For comparison, we repeated this experiment for the counterfactuals on only using the \abbrdiff ~(of Task 2) and TUMSyn by setting the disease label as a global condition. 

The TUMSyn requires paired brain scans from the same patient to learn how metadata influences image generation. We therefore selected patients with multiple MRIs, yielding 211 subjects and 614 scans, and constructed 1,662 intra-subject pairs. We then trained the approach for 40 epochs. 

As in the original publication, using the real MRIs revealed significant differences between the control and AUD group for the frontal lobe (p$\textless{0.0001}$), insula (p=$0.0008$), parietal lobe (p=$0.0002$), cingulate (p=$0.0004$), and temporal lobe (p$\textless{0.0001}$) but not for the occipital lobe (p=$0.9848$). Those findings were replicated when using the volumetric measurements extracted from the counterfactuals generated by our approach (Table~\ref{tab:ad_pvalues_combined}). This was not the case for \abbrdiff~(not significant for the cingulate and temporal lobes) and TUMSyn (not significant for the parietal, cingulate, and temporal lobes). More importantly, when repeating the analysis on just the controls by comparing their real values to the ones produced by their counterfactual AUD scores, only the measurements based on our approach can confirm the findings. This is also the case when confining analysis to the real AUD cases. Finally, when comparing the volume scores of real to counterfactual controls (i.e., from real AUD), no significant differences are correctly reported for ours and \abbrdiff, but significant differences are detected for TUM. The same is true when comparing real to counterfactual AUDs (i.e., generated from real controls).  In summary, only our method is able to produce counterfactuals that align with the original findings reported in \cite{sullivan2012using}.

%% file: pcgm_arxiv.bbl
\begin{thebibliography}{10}
\providecommand{\url}[1]{#1}
\csname url@samestyle\endcsname
\providecommand{\newblock}{\relax}
\providecommand{\bibinfo}[2]{#2}
\providecommand{\BIBentrySTDinterwordspacing}{\spaceskip=0pt\relax}
\providecommand{\BIBentryALTinterwordstretchfactor}{4}
\providecommand{\BIBentryALTinterwordspacing}{\spaceskip=\fontdimen2\font plus
\BIBentryALTinterwordstretchfactor\fontdimen3\font minus \fontdimen4\font\relax}
\providecommand{\BIBforeignlanguage}[2]{{%
\expandafter\ifx\csname l@#1\endcsname\relax
\typeout{** WARNING: IEEEtran.bst: No hyphenation pattern has been}%
\typeout{** loaded for the language `#1'. Using the pattern for}%
\typeout{** the default language instead.}%
\else
\language=\csname l@#1\endcsname
\fi
#2}}
\providecommand{\BIBdecl}{\relax}
\BIBdecl

\bibitem{mri_s1}
J.~Dubois, M.~Alison, S.~J. Counsell, L.~Hertz-Pannier, P.~S. H{\"u}ppi, and M.~J. Benders, ``{MRI} of the neonatal brain: a review of methodological challenges and neuroscientific advances,'' \emph{Journal of MRI}, vol.~53, no.~5, pp. 1318--1343, 2021.

\bibitem{mri_s2}
A.~Cagol \emph{et~al.}, ``{Association of brain atrophy with disease progression independent of relapse activity in patients with relapsing multiple sclerosis},'' \emph{JAMA neurology}, vol.~79, no.~7, pp. 682--692, 2022.

\bibitem{mri_s3}
D.~M. Sima \emph{et~al.}, ``{Artificial intelligence assistive software tool for automated detection and quantification of amyloid-related imaging abnormalities},'' \emph{JAMA Network Open}, vol.~7, no.~2, 2024, {Art. no.} e2355800.

\bibitem{GAN_gen}
A.~Sharma and G.~Hamarneh, ``Missing {MRI} {P}ulse {S}equence {S}ynthesis {U}sing {M}ulti-{M}odal {G}enerative {A}dversarial {N}etwork,'' \emph{IEEE Transactions on Medical Imaging}, vol.~39, no.~4, pp. 1170--1183, 2020.

\bibitem{mrigen}
P.-D. Tudosiu \emph{et~al.}, ``{Realistic morphology-preserving generative modelling of the brain},'' \emph{Nature Machine Intelligence}, vol.~6, no.~7, pp. 811--819, 2024.

\bibitem{jindal2024genAI}
J.~A. Jindal, M.~P. Lungren, and N.~H. Shah, ``{Ensuring useful adoption of generative artificial intelligence in healthcare},'' \emph{Journal of the American Medical Informatics Association}, vol.~31, no.~6, pp. 1441--1444, 2024.

\bibitem{wang2025self}
J.~Wang \emph{et~al.}, ``{Self-improving generative foundation model for synthetic medical image generation and clinical applications},'' \emph{Nature Medicine}, vol.~31, no.~2, pp. 609--617, 2025.

\bibitem{wang2025toward}
Y.~Wang \emph{et~al.}, ``{Toward general text-guided multimodal brain MRI synthesis for diagnosis and medical image analysis},'' \emph{Cell Reports Medicine}, vol.~6, no.~6, 2025, {Art. no.} 102182.

\bibitem{bluethgen2024vision}
C.~Bluethgen \emph{et~al.}, ``{A vision--language foundation model for the generation of realistic chest X-ray images},'' \emph{Nature Biomedical Engineering}, vol.~9, no.~4, pp. 494--506, 2024.

\bibitem{tudosiu2024realistic}
P.-D. Tudosiu \emph{et~al.}, ``{Realistic morphology-preserving generative modelling of the brain},'' \emph{Nature Machine Intelligence}, vol.~6, no.~7, pp. 811--819, 2024.

\bibitem{rombach2022ldm}
R.~Rombach, A.~Blattmann, D.~Lorenz, P.~Esser, and B.~Ommer, ``{High-resolution image synthesis with latent diffusion models},'' in \emph{2022 IEEE/CVF Conference on Computer Vision and Pattern Recognition (CVPR)}, 2022, pp. 10\,684--10\,695.

\bibitem{sd3}
P.~Esser \emph{et~al.}, ``{Scaling rectified flow transformers for high-resolution image synthesis},'' in \emph{Proceedings of the 41st International Conference on Machine Learning}, 2024, pp. 12\,606--12\,633.

\bibitem{dit}
W.~Peebles and S.~Xie, ``{Scalable diffusion models with transformers},'' in \emph{2023 IEEE/CVF International Conference on Computer Vision (ICCV)}, 2023, pp. 4195--4205.

\bibitem{wu2024eval}
J.~Wu, W.~Peng, B.~Li, Y.~Zhang, and K.~M. Pohl, ``{Evaluating the Quality of Brain MRI Generators},'' in \emph{{Medical Image Computing and Computer Assisted Intervention, Lecture Notes in Computer Science}}, vol. 15010, 2024, pp. 297--307.

\bibitem{singh2021medical}
N.~K. Singh and K.~Raza, ``{Medical image generation using generative adversarial networks: A review},'' in \emph{Health informatics: A computational perspective in healthcare}, 2021, pp. 77--96.

\bibitem{MCI}
C.~Ledig, A.~Schuh, R.~Guerrero, R.~A. Heckemann, and D.~Rueckert, ``{Structural brain imaging in Alzheimer’s disease and mild cognitive impairment: biomarker analysis and shared morphometry database},'' \emph{Scientific reports}, vol.~8,, no.~1, 2018, {Art. no.} 11258.

\bibitem{hivehsan}
J.~Zhang \emph{et~al.}, ``{Multi-label, multi-domain learning identifies compounding effects of HIV and cognitive impairment},'' \emph{Medical Image Analysis}, vol.~75, 2022, {Art. no.} 102246.

\bibitem{sullivan2018role}
E.~V. Sullivan \emph{et~al.}, ``{The role of aging, drug dependence, and hepatitis C comorbidity in alcoholism cortical compromise},'' \emph{JAMA Psychiatry}, vol.~75, no.~5, pp. 474--483, 2018.

\bibitem{sun2025eval}
\BIBentryALTinterwordspacing
P.~Sun, W.~Peng, L.~Li, Y.~Wang, and K.~M. Pohl, ``Evaluation of {3D} counterfactual brain {MRI} generation,'' \emph{Deep Generative Models, Lecture Notes in Computer Science}, Accepted. [Online]. Available: \url{https://arxiv.org/abs/2508.02880}
\BIBentrySTDinterwordspacing

\bibitem{cf_mri}
N.~J. Dhinagar, S.~I. Thomopoulos, E.~Laltoo, and P.~M. Thompson, ``{Counterfactual MRI Generation with Denoising Diffusion Models for Interpretable Alzheimer’s Disease Effect Detection},'' in \emph{IEEE Engineering in Medicine and Biology Society (EMBC)}, 2024, pp. 1--6.

\bibitem{puglisi2025brain}
L.~Puglisi, D.~C. Alexander, and D.~Rav{\`\i}, ``Brain latent progression: Individual-based spatiotemporal disease progression on {3D Brain MRIs} via latent diffusion,'' \emph{Medical Image Analysis}, vol. 106, 2025, {Art. no.} 103734.

\bibitem{fabio2023cfmodel}
F.~De~Sousa~Ribeiro, T.~Xia, M.~Monteiro, N.~Pawlowski, and B.~Glocker, ``High fidelity image counterfactuals with probabilistic causal models,'' in \emph{International Conference on Machine Learning}, vol. 202, 2023, pp. 7390--7425.

\bibitem{yeganeh2025latent}
Y.~Yeganeh \emph{et~al.}, ``{Latent Drifting in Diffusion Models for Counterfactual Medical Image Synthesis},'' in \emph{2025 IEEE/CVF Conference on Computer Vision and Pattern Recognition (CVPR)}, 2025, pp. 7685--7695.

\bibitem{peng2024latent}
\BIBentryALTinterwordspacing
W.~Peng \emph{et~al.}, ``{Latent Causal Modeling for 3D Brain MRI Counterfactuals},'' \emph{Deep Generative Models, Lecture Notes in Computer Science}, Accepted. [Online]. Available: \url{https://arxiv.org/abs/2409.05585}
\BIBentrySTDinterwordspacing

\bibitem{nestor2008ventricular}
S.~M. Nestor \emph{et~al.}, ``{Ventricular enlargement as a possible measure of Alzheimer's disease progression validated using the Alzheimer's disease neuroimaging initiative database},'' \emph{Brain}, vol. 131, no.~9, pp. 2443--2454, 2008.

\bibitem{controlnet}
L.~Zhang, A.~Rao, and M.~Agrawala, ``{Adding Conditional Control to Text-to-Image Diffusion Models},'' in \emph{2023 IEEE/CVF International Conference on Computer Vision (ICCV)}, 2023, pp. 3836--3847.

\bibitem{monaildm}
W.~H. Pinaya \emph{et~al.}, ``{Brain imaging generation with latent diffusion models},'' in \emph{Deep Generative Models, Lecture Notes in Computer Science}, vol. 13609, 2022, pp. 117--126.

\bibitem{ADNI}
R.~C. Petersen \emph{et~al.}, ``{Alzheimer's disease Neuroimaging Initiative (ADNI) clinical characterization},'' \emph{Neurology}, vol.~74, no.~3, pp. 201--209, 2010.

\bibitem{NCANDA}
S.~A. Brown \emph{et~al.}, ``{The National Consortium on Alcohol and NeuroDevelopment in Adolescence (NCANDA): a multisite study of adolescent development and substance use},'' \emph{Journal of studies on alcohol and drugs}, vol.~76, no.~6, pp. 895--908, 2015.

\bibitem{hivdata}
E.~Adeli, N.~M. Zahr, A.~Pfefferbaum, E.~V. Sullivan, and K.~M. Pohl, ``{Novel Machine Learning Identifies Brain Patterns Distinguishing Diagnostic Membership of Human Immunodeficiency Virus, Alcoholism, and Their Comorbidity of Individuals},'' \emph{Biological Psychiatry: Cognitive Neuroscience and Neuroimaging}, vol.~4, no.~6, pp. 589--599, 2019.

\bibitem{zitova2003image}
B.~Zitova and J.~Flusser, ``{Image registration methods: A survey},'' \emph{Image and vision computing}, vol.~21, no.~11, pp. 977--1000, 2003.

\bibitem{freeborough1996accurate}
P.~A. Freeborough, R.~P. Woods, and N.~C. Fox, ``Accurate registration of serial {3D} {MR} brain images and its application to visualizing change in neurodegenerative disorders,'' \emph{Journal of computer assisted tomography}, vol.~20, no.~6, pp. 1012--1022, 1996.

\bibitem{maintz1998survey}
J.~A. Maintz and M.~A. Viergever, ``{A survey of medical image registration},'' \emph{Medical image analysis}, vol.~2, no.~1, pp. 1--36, 1998.

\bibitem{hill2001medical}
D.~L. Hill, P.~G. Batchelor, M.~Holden, and D.~J. Hawkes, ``{Medical image registration},'' \emph{Physics in medicine \& biology}, vol.~46, no.~3, 2001, {Art. no.} R1.

\bibitem{chung2022score}
H.~Chung and J.~C. Ye, ``Score-based diffusion models for accelerated {MRI},'' \emph{Medical Image Analysis}, vol.~80, 2022, {Art. no.} 102479.

\bibitem{vae_mri}
A.~Volokitin \emph{et~al.}, ``{Modelling the Distribution of 3D Brain MRI Using a 2D Slice VAE},'' in \emph{Medical Image Computing and Computer Assisted Intervention, Lecture Notes in Computer Science}, vol. 12267, 2020, pp. 657--666.

\bibitem{MultiContrastGAN2019}
S.~U. Dar, M.~Yurt, L.~Karacan, A.~Erdem, E.~Erdem, and T.~Cukur, ``Image synthesis in multi-contrast {MRI} with conditional generative adversarial networks,'' \emph{IEEE Transactions on Medical Imaging}, vol.~38, no.~10, pp. 2375--2388, 2019.

\bibitem{shin2018medical}
H.-C. Shin \emph{et~al.}, ``{Medical image synthesis for data augmentation and anonymization using generative adversarial networks},'' in \emph{{Simulation and Synthesis in Medical Imaging, Lecture Notes in Computer Science}}, vol. 11037, 2018, pp. 1--11.

\bibitem{yu20183d}
B.~Yu, L.~Zhou, L.~Wang, J.~Fripp, and P.~Bourgeat, ``3{D} c{GAN} based cross-modality {MR} image synthesis for brain tumor segmentation,'' in \emph{2018 IEEE 15th International Symposium on Biomedical Imaging (ISBI 2018)}, 2018, pp. 626--630.

\bibitem{karras2019style}
T.~Karras, S.~Laine, and T.~Aila, ``A style-based generator architecture for generative adversarial networks,'' \emph{IEEE Transactions on Pattern Analysis and Machine Intelligence}, vol.~43, no.~12, pp. 4217--4228, 2021.

\bibitem{xing2021cycle}
S.~Xing, H.~Sinha, and S.~J. Hwang, ``Cycle consistent embedding of 3{D} brains with auto-encoding generative adversarial networks,'' in \emph{{Medical Imaging with Deep Learning}}, 2021, pp. 118--126.

\bibitem{bermudez2018learning}
C.~Bermudez, A.~J. Plassard, L.~T. Davis, A.~T. Newton, S.~M. Resnick, and B.~A. Landman, ``Learning implicit brain {MRI} manifolds with deep learning,'' in \emph{{Medical Imaging 2018: Image Processing}}, vol. 10574.\hskip 1em plus 0.5em minus 0.4em\relax SPIE, 2018, pp. 408--414.

\bibitem{han2018gan}
C.~Han \emph{et~al.}, ``{GAN}-based synthetic brain {MR} image generation,'' in \emph{2018 IEEE 15th International Symposium on Biomedical Imaging (ISBI 2018)}, 2018, pp. 734--738.

\bibitem{kwon2019generation}
G.~Kwon, C.~Han, and D.-s. Kim, ``{Generation of 3D Brain MRI Using Auto-Encoding Generative Adversarial Networks},'' in \emph{Medical Image Computing and Computer Assisted Intervention, Lecture Notes in Computer Science}, vol.~9, 2019, pp. 118--126.

\bibitem{ravi2022degenerative}
D.~Ravi \emph{et~al.}, ``{Degenerative adversarial neuroimage nets for brain scan simulations: Application in ageing and dementia},'' \emph{Medical image analysis}, vol.~75, 2022, {Art. no.} 102257.

\bibitem{peng2024brainsyn}
W.~Peng \emph{et~al.}, ``Metadata-conditioned generative models to synthesize anatomically-plausible {3D} brain {MRI}s,'' \emph{Medical Image Analysis}, vol.~98, 2024, {Art. no.} 103325.

\bibitem{ho2020ddpm}
J.~Ho, A.~Jain, and P.~Abbeel, ``{Denoising diffusion probabilistic models},'' in \emph{{Advances in Neural Information Processing Systems}}, vol.~33, 2020, pp. 6840--6851.

\bibitem{dhariwal2021diffusion}
P.~Dhariwal and A.~Nichol, ``Diffusion models beat {GAN}s on image synthesis,'' in \emph{{Advances in Neural Information Processing Systems}}, vol.~34, 2021, pp. 8780--8794.

\bibitem{la2022anatomically}
G.~La~Barbera \emph{et~al.}, ``Anatomically constrained {CT} image translation for heterogeneous blood vessel segmentation,'' in \emph{{British Machine Vision Virtual Conference}}, 2022, {Art. no.} 776.

\bibitem{wolleb2022diffusion}
J.~Wolleb, F.~Bieder, R.~Sandk{\"u}hler, and P.~C. Cattin, ``{Diffusion Models for Medical Anomaly Detection},'' in \emph{{Medical Image Computing and Computer-Assisted Intervention, Lecture Notes in Computer Science}}, vol. 13438, 2022, pp. 35--45.

\bibitem{dorjsembe2022three}
\BIBentryALTinterwordspacing
Z.~Dorjsembe, S.~Odonchimed, and F.~Xiao, ``Three-dimensional medical image synthesis with denoising diffusion probabilistic models,'' in \emph{Medical Imaging with Deep Learning}, 2022. [Online]. Available: \url{https://openreview.net/forum?id=Oz7lKWVh45H}
\BIBentrySTDinterwordspacing

\bibitem{peng2023cDPM}
W.~Peng, E.~Adeli, T.~Bosschieter, S.~H. Park, Q.~Zhao, and K.~M. Pohl, ``Generating realistic brain {MRIs} via a conditional diffusion probabilistic model,'' in \emph{{International Conference on Medical Image Computing and Computer-Assisted Intervention, Lecture Notes in Computer Science}}, vol. 14227, 2023, pp. 14--24.

\bibitem{yoon2023sadm}
J.~S. Yoon, C.~Zhang, H.-I. Suk, J.~Guo, and X.~Li, ``{Sadm: Sequence-aware diffusion model for longitudinal medical image generation},'' in \emph{{International Conference on Information Processing in Medical Imaging}}, 2023, pp. 388--400.

\bibitem{han2023medgen3d}
K.~Han \emph{et~al.}, ``Med{Gen3D}: A deep generative framework for paired {3D} image and mask generation,'' in \emph{Medical Image Computing and Computer Assisted Intervention, Lecture Notes in Computer Science}, vol. 14220, 2023, pp. 759--769.

\bibitem{pombo2023equitable}
G.~Pombo \emph{et~al.}, ``Equitable modelling of brain imaging by counterfactual augmentation with morphologically constrained {3D} deep generative models,'' \emph{Medical Image Analysis}, vol.~84, 2023, {Art. no.} 102723.

\bibitem{thiagarajan2022training}
J.~J. Thiagarajan, K.~Thopalli, D.~Rajan, and P.~Turaga, ``{Training calibration-based counterfactual explainers for deep learning models in medical image analysis},'' \emph{Scientific reports}, vol.~12, no.~1, 2022, {Art. no.} 597.

\bibitem{billot_robust_2023}
B.~Billot, Y.~Colin, Magdamo~Cheng, S.~Das, and J.~E. Iglesias, ``{Robust} machine learning segmentation for large-scale analysis of heterogeneous clinical brain {MRI} datasets,'' \emph{{Proceedings} of the {National} {Academy} of {Sciences} ({PNAS})}, vol. 120, no.~9, 2023, {Art. no.} e2216399120.

\bibitem{pawlowski2020deep}
N.~Pawlowski, D.~Coelho~de Castro, and B.~Glocker, ``Deep structural causal models for tractable counterfactual inference,'' \emph{{Advances in Neural Information Processing Systems}}, vol.~33, pp. 857--869, 2020.

\bibitem{rezende2015variational}
D.~Rezende and S.~Mohamed, ``{Variational inference with normalizing flows},'' in \emph{{International Conference on Machine Learning}}, vol.~37, 2015, pp. 1530--1538.

\bibitem{de2023high}
F.~De~Sousa~Ribeiro, T.~Xia, M.~Monteiro, N.~Pawlowski, and B.~Glocker, ``High fidelity image counterfactuals with probabilistic causal models,'' in \emph{International Conference on Machine Learning}, 2023, pp. 7390--7425.

\bibitem{jernigan1991reduced}
T.~L. Jernigan \emph{et~al.}, ``{Reduced cerebral grey matter observed in alcoholics using magnetic resonance imaging},'' \emph{Alcoholism: Clinical and Experimental Research}, vol.~15, no.~3, pp. 418--427, 1991.

\bibitem{causalvae}
\BIBentryALTinterwordspacing
L.~Chen \emph{et~al.}, ``Od-vae: An omni-dimensional video compressor for improving latent video diffusion model,'' \emph{arXiv preprint arXiv:2409.01199}, 2024. [Online]. Available: \url{https://arxiv.org/abs/2409.01199}
\BIBentrySTDinterwordspacing

\bibitem{bain2022frozentimejointvideo}
M.~Bain, A.~Nagrani, G.~Varol, and A.~Zisserman, ``{Frozen in time: A joint video and image encoder for end-to-end retrieval},'' in \emph{2021 IEEE/CVF Conference on Computer Vision and Pattern Recognition (CVPR)}, 2021, pp. 1728--1738.

\bibitem{xu2024medsyn}
Y.~Xu \emph{et~al.}, ``{MedSyn}: Text-guided anatomy-aware synthesis of high-fidelity {3D} {CT} images,'' \emph{IEEE Transactions on Medical Imaging}, vol.~43, no.~10, pp. 3648--3660, 2024.

\bibitem{zhang2023adding}
L.~Zhang, A.~Rao, and M.~Agrawala, ``{Adding conditional control to text-to-image diffusion models},'' in \emph{2023 IEEE/CVF International Conference on Computer Vision (ICCV)}, 2023, pp. 3836--3847.

\bibitem{diffvae}
K.~Preechakul, N.~Chatthee, S.~Wizadwongsa, and S.~Suwajanakorn, ``Diffusion autoencoders: Toward a meaningful and decodable representation,'' in \emph{2022 IEEE/CVF Conference on Computer Vision and Pattern Recognition (CVPR)}, 2022, pp. 10\,609--10\,619.

\bibitem{Kingma2013VAE}
\BIBentryALTinterwordspacing
D.~P. Kingma and M.~Welling, ``{Auto-Encoding Variational Bayes},'' in \emph{2nd International Conference on Learning Representations}, Y.~Bengio and Y.~LeCun, Eds., 2014. [Online]. Available: \url{http://arxiv.org/abs/1312.6114}
\BIBentrySTDinterwordspacing

\bibitem{mokady2022inv}
R.~Mokady, A.~Hertz, K.~Aberman, Y.~Pritch, and D.~Cohen-Or, ``Null-text inversion for editing real images using guided diffusion models,'' in \emph{2023 IEEE/CVF Conference on Computer Vision and Pattern Recognition (CVPR)}, 2023, pp. 6038--6047.

\bibitem{vqvae2017}
A.~van~den Oord, O.~Vinyals, and K.~Kavukcuoglu, ``{Neural Discrete Representation Learning},'' in \emph{{Advances in Neural Information Processing Systems}}, 2017, pp. 6309--6318.

\bibitem{ferreira2022alphaGAN}
A.~Ferreira, R.~Magalh{\~a}es, S.~M{\'e}riaux, and V.~Alves, ``{Generation of Synthetic Rat Brain MRI Scans with a 3D Enhanced Alpha Generative Adversarial Network},'' \emph{Applied Sciences}, vol.~12, no.~10, 2022, {Art. no.}, 4844.

\bibitem{sun2022hagan}
L.~Sun, J.~Chen, Y.~Xu, M.~Gong, K.~Yu, and K.~Batmanghelich, ``Hierarchical amortized {GAN} for {3D} high resolution medical image synthesis,'' \emph{IEEE Journal of Biomedical and Health Informatics}, vol.~26, no.~8, pp. 3966--3975, 2022.

\bibitem{pinaya2022brain}
W.~H. Pinaya \emph{et~al.}, ``{Brain imaging generation with latent diffusion models},'' in \emph{Deep Generative Models, Lecture Notes in Computer Science}, vol. 13609, 2022, pp. 117--126.

\bibitem{petersen2010alzheimer}
R.~C. Petersen \emph{et~al.}, ``{Alzheimer's Disease Neuroimaging Initiative} ({ADNI}): clinical characterization,'' \emph{Neurology}, vol.~74, no.~3, pp. 201--209, 2010.

\bibitem{cfg}
\BIBentryALTinterwordspacing
J.~Ho and T.~Salimans, ``{Classifier-Free Diffusion Guidance},'' in \emph{{NeurIPS 2021 Workshop on Deep Generative Models and Downstream Applications}}, 2021. [Online]. Available: \url{https://openreview.net/forum?id=qw8AKxfYbI}
\BIBentrySTDinterwordspacing

\bibitem{fid}
M.~Heusel, H.~Ramsauer, T.~Unterthiner, B.~Nessler, and S.~Hochreiter, ``{GANs trained by a two time-scale update rule converge to a local {Nash} equilibrium},'' \emph{{Advances in Neural Information Processing Systems}}, vol.~30, pp. 6626--6637, 2017.

\bibitem{msssim}
Z.~Wang, E.~Simoncelli, and A.~Bovik, ``{Multiscale structural similarity for image quality assessment},'' in \emph{The Thrity-Seventh Asilomar Conference on Signals, Systems \& Computers, 2003}, vol.~2, 2003, pp. 1398--1402.

\bibitem{freesurfer}
B.~Fischl, ``{FreeSurfer},'' \emph{NeuroImage}, vol.~62, no.~2, pp. 774--781, 2012.

\bibitem{abcd}
B.~J. Casey \emph{et~al.}, ``{The Adolescent Brain Cognitive Development (ABCD) study: Imaging acquisition across 21 sites},'' \emph{Developmental cognitive neuroscience}, vol.~32, pp. 43--54, 2018.

\bibitem{sudlow2015ukbiobank}
C.~Sudlow \emph{et~al.}, ``{UK Biobank: An open access resource for identifying the causes of a wide range of complex diseases of middle and old age},'' \emph{PLoS medicine}, vol.~12, no.~3, 2015, {Art. no.} e1001779.

\bibitem{van2012HCP}
D.~C. Van~Essen \emph{et~al.}, ``{The Human Connectome Project: A data acquisition perspective},'' \emph{Neuroimage}, vol.~62, no.~4, pp. 2222--2231, 2012.

\bibitem{sullivan2012using}
G.~M. Sullivan and R.~Feinn, ``{Using effect size—or why the P value is not enough},'' \emph{{Journal of Graduate Medical Education}}, vol.~4, no.~3, pp. 279--282, 2012.

\end{thebibliography}
